\documentclass[10pt,twocolumn,letterpaper]{article}

\usepackage{wacv}

\usepackage[utf8]{inputenc}
\usepackage[T1]{fontenc}

\usepackage{times}
\usepackage{epsfig}
\usepackage{graphicx}
\usepackage{amsmath}
\usepackage{amssymb}

\usepackage{array}
\usepackage{booktabs}
\usepackage{amsfonts}
\usepackage{siunitx}
\usepackage{enumitem}
\usepackage[caption=false]{subfig}
\usepackage{hyperref}

\wacvfinalcopy 

\def\wacvPaperID{****} 

\ifwacvfinal\fi
\begin{document}

\title{Attention Flow: End-to-End Joint Attention Estimation}


\author{Ömer Sümer$^1$, Peter Gerjets$^2$, Ulrich Trautwein$^1$, Enkelejda Kasneci$^1$\\
$^1$ University of Tübingen \quad  $^2$ Leibniz-Institut für Wissensmedien \\
Tübingen, Germany\\
{\tt\small \{oemer.suemer,ulrich.trautwein,enkelejda.kasneci\}@uni-tuebingen.de}\\
{\tt\small p.gerjets@iwm-tuebingen.de}
}

\maketitle
\ifwacvfinal\thispagestyle{empty}\fi

\begin{abstract}
    This paper addresses the problem of understanding joint attention in third-person social scene videos. Joint attention is the shared gaze behaviour of two or more individuals on an object or an area of interest and has a wide range of applications such as human-computer interaction, educational assessment, treatment of patients with attention disorders, and many more. Our method, Attention Flow, learns joint attention in an end-to-end fashion by using saliency-augmented attention maps and two novel convolutional attention mechanisms that determine to select relevant features and improve joint attention localization. We compare the effect of saliency maps and attention mechanisms and report quantitative and qualitative results on the detection and localization of joint attention in the VideoCoAtt dataset, which contains complex social scenes.
\end{abstract}

\section{Introduction}
Humans spend most of their lives interacting with each other. In public or private spaces such as squares, concert halls, cafes, schools, we share various aspects of everyday life with one another. Through new technologies and growing distractive effects of social media, we divide our attention and memory into separate themes and may have difficulties to focus our attention onto our primary task. In that regard, from both psychological and computer vision perspectives, understanding a person's attentional focus and particular localization of joint attention present valuable opportunities. 

\begin{figure}[ht!]
	\centering
	\subfloat[input image]
	{\includegraphics[width=.205\textwidth]{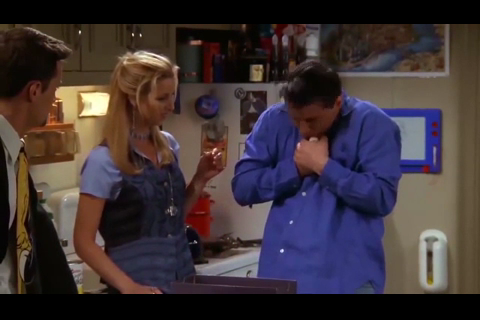}}\quad
	\subfloat[saliency estimation]
	{\includegraphics[width=.205\textwidth]{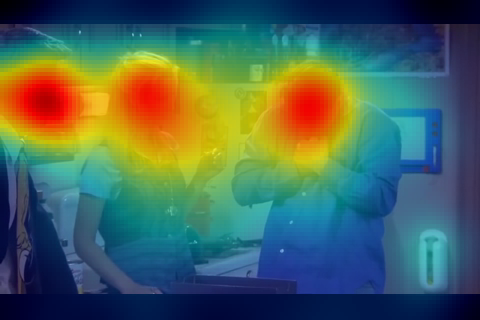}}\\
	\subfloat[face likelihood]
	{\includegraphics[width=.205\textwidth]{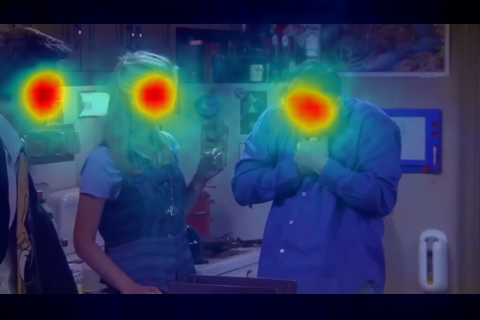}}\quad
	\subfloat[co-attention likelihood]
	{\includegraphics[width=.205\textwidth]{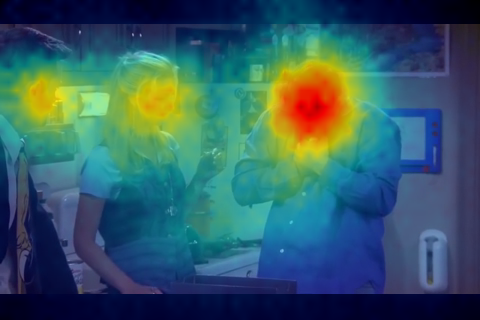}}
	\caption{Sample of a social scene in (a), and the estimated saliency map using \cite{Kummerer:2016} in (b). Our method, \emph{Attention Flow} takes only the input image in (a) and estimate the face likelihood (c) and  the co-attention likelihood (d).}
	\label{figure:01}
\end{figure}

Joint attention is very helpful in many different contexts. For example, in classroom-based learning, teachers who engage all students equally can enhance student achievement \cite{Goldberg:2019,Kent:2013,Prieto:2018,Watanabe:2013}. To investigate this, educational researchers manually analyze student behaviours and especially the visual attention of students from video recordings of instructions and try to explain relationships between students' and teachers' behaviour in a very time-consuming way. Another example is in the context of attention disorders or autism research. For instance, it has been shown that joint attention and engagement, particularly in early ages, can be taught using behavioural and developmental interventions \cite{Olswang:2014}. Thus, computer vision-based, automated joint attention analysis can be instrumental in behavioural psychology to develop efficient training curricula for the treatment of children with disabilities. Another useful application is in the area of human-computer interaction and especially interaction with autonomous systems. For example, robots can infer gaze direction in case of a single person or joint attention in groups and turn their heads into that direction. Such information could be further used by robots to augment their collaboration with humans \cite{Andrist:2015, Aronson:2018}.

Although an automated analysis of joint attention might be beneficial for a variety of applications, related work in the domain of computer vision is still quite limited. Few works addressed a similar problem, namely social saliency in first and third-person view \cite{Park:2012, Park:2013, Park:2015, Suemer:2018}. Also, there are examples of joint attention in human-robot interaction \cite{Andrist:2015,Silva:2012,Nagai:2006,Nagai:2003}. Whereas mapping gaze directions to a common plane \cite{Santini:2017} is a promising option in controlled settings, it does not work in more challenging multimedia data. A recent study \cite{Fan:2018} collected a large video dataset, which we use in this study, and proposed a spatiotemporal neural network to estimate shared attention. Even though we deal with the same problem, we prefer to use the term of joint attention, since shared attention, from a psychological perspective, includes further underlying cognitive processes and does not necessitate joint gaze.

In this work, we propose a new approach that relates saliency and joint attention to estimate locations of joint attention in third person images or videos. Simply explained, saliency is an estimation of fixation likelihood on an image. In fact, due to the limited capacity of our visual system, we, by the help of an attentional mechanism, focus on the most relevant parts of a scene that are more distinctive than the remaining. In essence, it is how our eye movements process a scene, by employing various eye movements (such as saccade and fixations) and visual search which is guided by various bottom-up and top-down processes. Eye tracking-based saliency information has supported many computer vision tasks such as object detection \cite{Papadopoulos:2014}, zero-shot image classification \cite{Karessli:2017}, and image/video captioning \cite{Cornia:2018,Yu:2017}.

Figure~\ref{figure:01} shows a sample of our approach. Despite the usefulness of saliency maps, they do not necessarily represent the visual focus of people in the scene. However, during the training time, we exploit saliency maps to encode contextual information and create pseudo attention maps by combining them with face locations and their joint attention point and learn to predict these likelihoods. Then, during the test time, we can summarise the attentional focus of people in given third-person social images or videos.  
 
The main contributions of this paper are as follows:
\setlist{nolistsep}
\begin{enumerate}[noitemsep]
	\item It formulates the problem of inferring joint attention as end-to-end training. Thereby, \textit{Attention Flow} works without additional dependencies such as face/head detection, region proposals, or saliency estimation.
	\item It explicitly learns saliency and joint attention of a high-level inference task using saliency augmented pseudo attention maps and Attention Flow network with channel-wise and spatial attention mechanisms.
	\item Experimental results verify the performance of our approach on large-scale social videos, namely the VideoCoAtt dataset \cite{Fan:2018}. We also present a comparative ablation analysis of saliency and attention modules.  
\end{enumerate}     

\section{Related Work}

First, we review related research on gaze following and joint attention. Then, we will discuss saliency estimation and attention modules as we utilized them in our approach to infer the joint attention. \\
\indent \textbf{Gaze Following:\ }Recasens \emph{et al.} \cite{Recasens:2015} proposed a neural network which predicts the locations being gazed at in a convolutional neural network using head location, an image patch from head location, and an entire image. They also created a large-scale dataset where persons' eye and gaze locations were annotated. They later extended their work to use eye locations in a video frame and to predict gazed location in future frames \cite{Recasens:2017}. Gorji \emph{et al.} \cite{Gorji:2017} used a similar approach to \cite{Recasens:2015}; however, they leveraged gaze information to boost saliency estimation and did not report gaze following results. 

Recently, Chong \emph{et al.} \cite{Chong:2018} proposed a method to train gaze following, head pose and gaze tasks based on a multitask learning approach by optimizing several losses on different tasks and datasets. They also included \emph{outside} of the frame labels and predicted visual attention. Nevertheless, their approach estimates a single person's visual attention, not joint attention.\\
\indent \textbf{Joint vs. Shared Attention:\ } Joint attention is a social interaction that can occur in the forms of dyadic (looking at each other) or triadic ways (looking at each other and an object). Previous research shows that infants can discriminate between dyadic and joint attention interactions already by the age of 3 months \cite{Striano:2006}. Joint attention is crucial for language learning and imitative learning \cite{Baldwin:1993,Murray:2008}. In contrast to joint attention, shared attention does not require co-attending physically or by gaze. For instance, co-attending a television broadcast when looking at another point can be an example of shared attention. An observer can understand shared attention by using cues from the environment \cite{Shteynberg:2015}. Shared attention is more related to the underlying cognitive processes, whereas joint attention is dyadic and triadic gaze oriented. Thus, in the following we will use the term of joint attention since computer vision relies on seen visual cues.\\
\begin{figure*}[ht!]
	\centering
	\includegraphics[width=\linewidth]{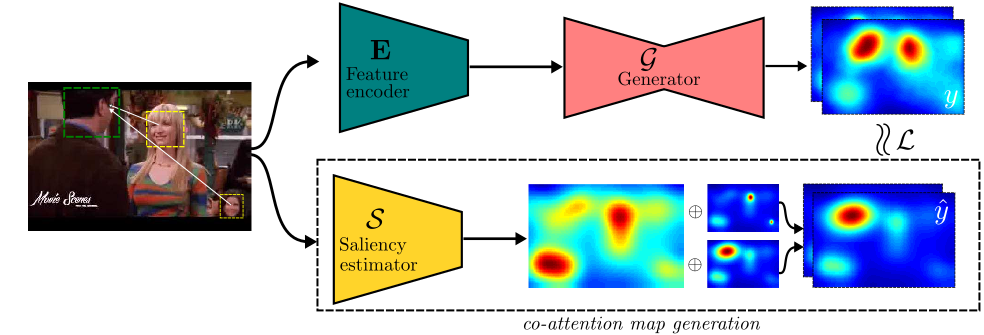}
	\caption{\emph{Overview of our Attention Flow} Our method is composed of three modules, (i) feature encoder, (ii) attention flow generator, and (iii) saliency-based ground truth generation. It estimates a two-channel heatmap, which encodes faces and their co-attention likelihood in the scene.}
	\label{figure:02}
\end{figure*}
\indent \textbf{Joint Attention:\ }Looking into studies on the analysis of attention in social interactions, \cite{Park:2012} localized head-mounted cameras in 3D using structure from motion and triangulated joint attention. Later, they proposed a geometric model between joint attention and social formation captured from first and third person views \cite{Park:2015}. These works are noteworthy; however, they depend on first-person views and thus cannot be applied in unconstrained third view images and videos. Also, they aim to predict only proximity of joint attention (social saliency) and cannot present a good understanding of joint attention.\\
\indent \textbf{Saliency Estimation:\ }Saliency is a measure of spatial importance, and it characterizes the parts of the scene which stand out relative to other parts. Being salient can depend on low-level features such as luminance, color, texture, high-level features such as objectness, task-driven factor, and center bias phenomenon. In the literature of saliency estimation, two approaches exist: (a) bottom-up methods, which aim to combine relevant information without prior knowledge of the scene, and (b) top-down methods which are more goal-oriented \cite{Borji:2013}. Availability of large-scale attention datasets and deep learning approaches have surpassed all previous psychological and computational methods. Based on these recent studies, we know that humans look at humans, faces, objects, texts \cite{Bylinskii:2016} and also emotional content \cite{Fan_2018_emotional}. The joint attention of humans in the scene is also noticeable. For this reason, we will leverage saliency information to learn joint attention.\\
\indent \textbf{Attention Mechanism:\ } Computer-based estimation of attention can also be approached by means of machine-learning techniques, where models, with the help of spatial or temporal attention mechanisms, are able to learn where, when, and what to attend. The use of  First use cases are machine translation \cite{Bahdanau:2014}, image captioning \cite{Kelvin:2015}, and action classification \cite{Sharma:2015}.\\
\indent Looking into attention mechanisms in images, Wang \emph{et al.} \cite{Wang:2017} incorporated attention modules into an encoder-decoder network and performed well in an image classification task. Their method learns attention jointly in 3D. Another recent work exploited inter-channel relationships. In Squeeze-and-Excitation blocks, they utilized global average-pooled features to perform a channel-wise calibration \cite{Hu:2018}. Recently, Woo et al. \cite{Woo:2018} proposed a convolutional attention module that leverages channel and spatial relations separately.

The common point of these works is that they address classification tasks by the use of spatial, temporal, or channel-wise attention. In contrast, we propose novel convolutional attention mechanisms for two purposes: the first is to learn feature selection along the channel dimension of a learned representation, and secondly, to guide a regression network to focus on more relevant areas in the spatial dimension. Instead of an architectural block in a classification task as in \cite{Hu:2018}, we utilize these blocks to benefit from learned features better by applying an adaptive feature selection and apply a further refinement on top of the heatmap generation module. 

\section{Method}
Our approach aims to infer joint attention in third person social videos, where two or more people look at another person or object. Figure \ref{figure:02} shows an overview of our workflow.

For a given social image or video frame, we estimate a two-channel likelihood distribution, called \emph{Attention Flow}. One channel represents faces in the scene, whereas the second channel is the likelihood of joint attention. In our workflow, raw images can be considered as 
a fusion of social presence in the scene and the center of joint attention. Figure \ref{figure:02} depicts an example prediction of our approach. Our Attention Flow network takes only raw images and detects faces and their respective co-attention locations without depending on any other information. In this section, we will describe (1) the creation of pseudo-attention maps (\S \ref{ssec:pseudoattention}), which are augmented by saliency estimation; (2) learning and inference (\S \ref{ssec:learning}) by our Attention Flow network using attention mechanisms, and provide (3) implementation details  (\S \ref{ssec:implementation}).

\subsection{Saliency Augmented Pseudo-Attention Maps}\label{ssec:pseudoattention}
Consider persons interacting with each other in a social scene. The question we address is how to infer their visual attention focus from a third person's view? Probably the most accurate way to obtain this information would be by employing mobile eye trackers or through gaze estimation based on several high-resolution field cameras. For the majority of use cases in our daily lives, where such equipment cannot be employed, it would be very useful to be able to retrieve such information solely based on images or video material. For this reason, we first compute pseudo-attention maps by leveraging saliency estimation. \\

More specifically, for an input image $I$, we have a number of detected head locations $(x_i, y_i, w_i, h_i)$, where $i=1,2,...,n$ and $n\geq 0$. To model social presence and the respective co-attention location we use Gaussian distributions. For a head detection or co-attention bounding box, this distribution is defined as

\begin{equation}
\begin{aligned}
\mathcal{G}&(x+\delta x,y+\delta y)= \\
& \begin{cases}
\exp \{ -\frac{x^2+y^2}{2\sigma^2} \} \quad \ \Vert \delta x\Vert\leq w, \Vert \delta y\Vert\leq h\\
0, \quad otherwise
\end{cases}
\end{aligned}
\end{equation}

Then, we combine head locations and co-attention maps with the estimated saliency maps, which is a precursor of observer's attention. Augmented by estimated saliency maps $\mathcal{S}$, the created pseudo-attention maps can be formalized as follows:  

\begin{equation}\label{eq:attention_map}
\begin{aligned}
& \mathcal{H}_1 = \alpha \log \Big(\sum_{i=1,..,n}\mathcal{G}_{f_i}\Big) + \beta \log(\mathcal{S})  \\
&\mathcal{H}_2 =  \alpha \log(\mathcal{G}_{coatt}) + \beta \log(\mathcal{S}) 
\end{aligned}
\end{equation}

In this way, we suppress the saliency to lower values and ensure that if there are detected faces in the scene, they and their respective co-attention point will correspond to the maximum values of pseudo-attention maps in the first and second channels.

By employing saliency estimation in our method, we leverage the information of relative importance of the regions which can be also salient for the persons in the scene. Thus, it prevents unreliable training samples, where the same object can appear as a co-attention point or zero when we use only $\mathcal{G}_{f}$ and $\mathcal{G}_{coatt}$.     

\subsection{Attention Flow Network}\label{ssec:learning}

Our model aims to solve three problems simultaneously: (1) to locate faces in the given image, (2) to detect whether joint attention exists or not, and (3) to predict the location of joint attention.

As input we only use the raw images instead of any other computational blocks, such as face detector, object detector or proposal networks. In this way, our Attention Flow network can be used to retrieve images or videos according to their social context in an efficient and fast way. The two-channel saliency augmented pseudo-attention maps are a compressed form of these objectives and provide all necessary information. In images which do not contain faces, the first channel of the attention map will give a lower likelihood, and they can be easily omitted from the further attention analysis.

In case of two or more persons in the scene, the first channel will represent the locations of their faces, whereas the second channel will be either estimated saliency or joint attention. Since pseudo-attention maps are a weighted summation of saliency estimation and joint attention, the typical values of maximum points are informative about the presence of joint attention. Therefore, learning pseudo-attention maps enables both detection and localization tasks simultaneously.

As it can be seen in Figure \ref{figure:02}, we first extract a visual representation of the scene using a pre-trained encoder network on object classification tasks. Since inferring joint attention is a complex problem even for humans, we leverage from an encoder to understand the visual focus of the persons in the image and for better generalization. The following block is a generator network, which learns attention maps from encoded representations. In order to avoid undesired outcomes of rescaling, we preserve the original aspect ratio in the input image and prefer fully convolutional architectures in both encoder and generator networks.

As a loss function, we use the Mean Squared Error (MSE) between the predicted attentions maps $\hat{\mathcal{H}}$ and ground truth pseudo saliency maps $\mathcal{H}$ (created as described in \S \ref{ssec:pseudoattention}):

\begin{equation}
\mathcal{L}_{MSE}= \frac{1}{H \cdot W} \Vert G(E(I)), \mathcal{H} \Vert^2
\end{equation}

\begin{figure}[ht!]
	\centering
	\includegraphics[width=\linewidth]{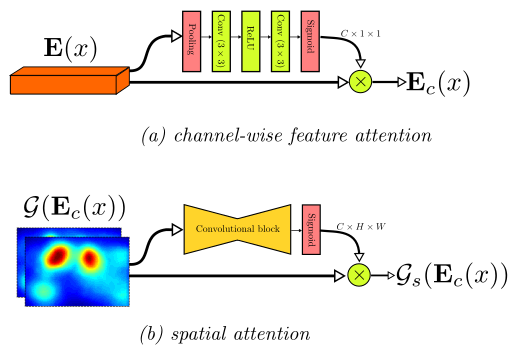}
	\caption{Channel-wise feature attention and convolutional spatial attention blocks.}
	\label{figure:03}
\end{figure}

When compared to other vision tasks such as object detection, segmentation or categorization, localizing the joint attention is a very complex task because the same region, i.e., a face or an object, can be the co-attention point in a scene, but shortly for a short period of time, it might not be true. In order to deal with these situations, our Attention Flow network can be guided towards the more relevant regions. For this purpose, we propose two novel attention mechanisms, namely \emph{channel-wise} and \emph{spatial}, and investigate their efficiency in the localization of joint attention. Figure \ref{figure:03} shows these attention mechanisms.

The encoder output is typically the output of a convolutional network which preserves spatial information in a reduced resolution and contains a higher dimension in the channel. The combination of these feature maps decides whether objects are present in the image. Using the complete encoded representation is redundant. According to the context, some channels can have more importance in the representation of the scene. Channel-wise convolutional attention performs a feature selection by weighting channels according to their contribution to the task.

On the other hand, spatial attention works as a refinement on top of the final joint attention estimations. In contrast to the spatial attention mechanisms in classification, which works as an importance map to maximize class activations, our spatial attention augments a heatmap regression task.

\subsection{Implementation Details}\label{ssec:implementation}
The Attention Flow network is composed of three main modules: encoder, generator, and co-attention map generation blocks. In order to exploit the knowledge of large-scale object classification tasks, we use a pre-trained ResNet-50 \cite{He:2016} as an encoder. Our final estimation is an attention map and needs to preserve spatial relations as much as possible. Thus, we prefer dilated residual architecture, DRN-A-50 \cite{Yu:2017_dilated} trained on ImageNet and keep the resolution 1/8 at the output of the encoder. 

As a generator, we used 9 residual blocks with instance normalization. It takes inputs in the number of feature channels (2048) and outputs 2-channel attention maps. Then, linear upsampling (x8) is applied. 

The last block is co-attention map generation, and it is used only in training time to produce ground truth attention maps as described in \S \ref{ssec:pseudoattention}. To estimate saliency, we used Deep Gaze II \cite{Kummerer:2016}. Similar to other data-driven saliency estimation methods, Deep Gaze II makes use of different level of features and has an understanding of objectness. It helps us to reduce the number of potential locations where joint attention might exist.

The layers of channel-wise feature attention are depicted in Figure~\ref{figure:03}(a). On the other hand, in the convolutional block of spatial attention, we used a small residual network that contains 3 residual blocks. As we applied spatial attention at 1/8 resolution before upsampling, it does not introduce an extensive computational cost to the entire workflow.

At training time, we used a SGD solver with a learning rate of 0.01 in the generator block. In feature encoder, we either lock the pre-trained parameters or applied \emph{fine-tuning} by a 10 times reduced learning rate.

\section{Experiments}

In this section, we first define the used dataset and performance metrics. Then, we report the ablation studies on the use of saliency estimation to create attention maps and the effect of attention mechanisms and evaluate our approach on the VideoCoAtt dataset in comparison to related approaches.

\subsection{Experimental Setup}\label{ssec:experimentalsetup}

To evaluate our approach on joint attention estimation, we used the Video Co-Attention dataset \cite{Fan:2018}, which is currently, to the best of our knowledge, the only available dataset on a joint attention task. The dataset contains 380 RGB video sequences from 20 different TV shows in the resolution of 320$\times$480. There are 250,030 frames in the training set, 128,260 frames in the validation set, and 113,810 frames in the testing set. Each split comes from different TV shows, and the dataset includes varying human appearances and formation.

There are two tasks: detection and localization of joint attention. Some images might not contain human bodies or faces. In images with social content, subjects' attentional focus can be different. In the detection task, we report overall prediction accuracy in the test set of VideoCoAtt. On the other hand, localization is evaluated on the test images with joint attention locations. $L_2$ distance in the input resolution will be used.

By adopting the evaluation procedure from \cite{Fan:2018}, we use the Structured Edge Detection Toolbox \cite{Zitnick:2014} to generate bounding box proposals. In the location, where our method predicts joint attention, we apply a Non-Maximum Suppression (NMS) and take the one that intersects greatest with our predicted estimation. It should be noted that our approach can locate the center of joint attention. Thus, in order to make a fair comparison with state-of-the-art methods, and we used the bounding box proposal.

Furthermore, there may be no joint attention or more than one joint attention location in an image. In order to learn the detection and localization of joint attention at the same time, we learn by all types of images without social context (body or faces), with social context but without any joint attention, and one or more joint attention. According to Eq. \ref{eq:attention_map}, we limit the values of saliency to some range by natural logarithm and a scale factor. Thus, our trained network's prediction can be joint attention if and only if the predicted likelihood is greater than a threshold.   

\begin{figure}[t]
	\centering
	\includegraphics[width=\linewidth]{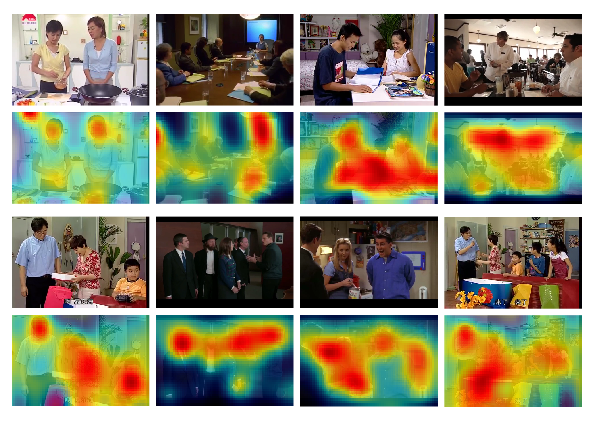}
	\caption{Example daily life scenes from VideoCoAtt dataset \cite{Fan:2018} and their respective saliency estimations using Deep Gaze II \cite{Kummerer:2016}. The focus of shared visual attention does not necessarily need to be the most salient region, but contains auxiliary information to localize joint attention.}
	\label{figure:04}
	\vspace{-0.40cm}
\end{figure}

\subsection{Results and Analysis}\label{ssec:ablation}
\paragraph*{Saliency and joint attention} Saliency models the attention of a third person who observes a video or image. On the other hand, joint attention analysis aims to understand from the perspectives of persons in these visual content. Due to the geometric difference between the viewpoints and human behavior in social scenes, the most salient part of images may not be the focus of persons' attention. Thus, we investigate how the co-attention locations are salient for different saliency estimation methods.

We tested four saliency estimation methods, Itti and Koch \cite{Itti:1998}, GBVS \cite{Harel:2007}, Signature \cite{Hou:2012}, and Deep Gaze 2 \cite{Kummerer:2016}. The first three were chosen as representatives of classical computational saliency methods, whereas Deep Gaze 2 is a data-driven approach that depends on pre-trained feature representations on image classification. Deep Gaze 2's mean saliency value in co-attention bounding boxes of the training images, 96\% of the time, are above the mean saliency value of images, whereas it is the case in 44\%, 71\% and 77\% for Itti \& Koch, GVBS, and Signature.

In most cases, persons in the scene interact with either another person or an object. We regard that a data-driven Deep Gaze 2 can result in higher saliency in co-attention regions as it leverages a representation trained on object classification. Thus, we prefer Deep Gaze 2 when creating pseudo attention maps (\S \ref{ssec:pseudoattention}). 

Figure \ref{figure:04} shows some sample images and their estimated saliency maps using Deep Gaze 2 \cite{Kummerer:2016}, respectively. These samples show us that the most salienct regions do not necessarily contain the possible joint attention in the social images. However, they are a precursor of observer's attention who gaze at images.

A tiny visual change in the image can cause a big change in the presence and location of joint attention. This is the main reason why we leverage the saliency information. The ``raw image'' results in \cite{Fan:2018} also validate our assumption. One can suppose the use of saliency as introducing noise, however, starting from the attention of observer and guiding the attention of the network towards understanding the attention of people inside the scene is a reasonable solution and also makes the problem learnable.
 
\vspace{-0.2cm}
\paragraph*{Use of attention mechanisms} Our Attention Flow network learns joint attention by using a pre-trained representation and a generator as a regression task with mean square loss. To supplement it, we proposed two attention mechanisms. In contrast to existing attention mechanisms in the literature, such as temporal in videos or text data, or spatial in image categorization, we use two novel convolutional attention blocks for feature selection and regression tasks. We evaluate their performance on the joint attention localization task. 

\begin{figure*}[ht!]
	\centering
	\includegraphics[width=\linewidth]{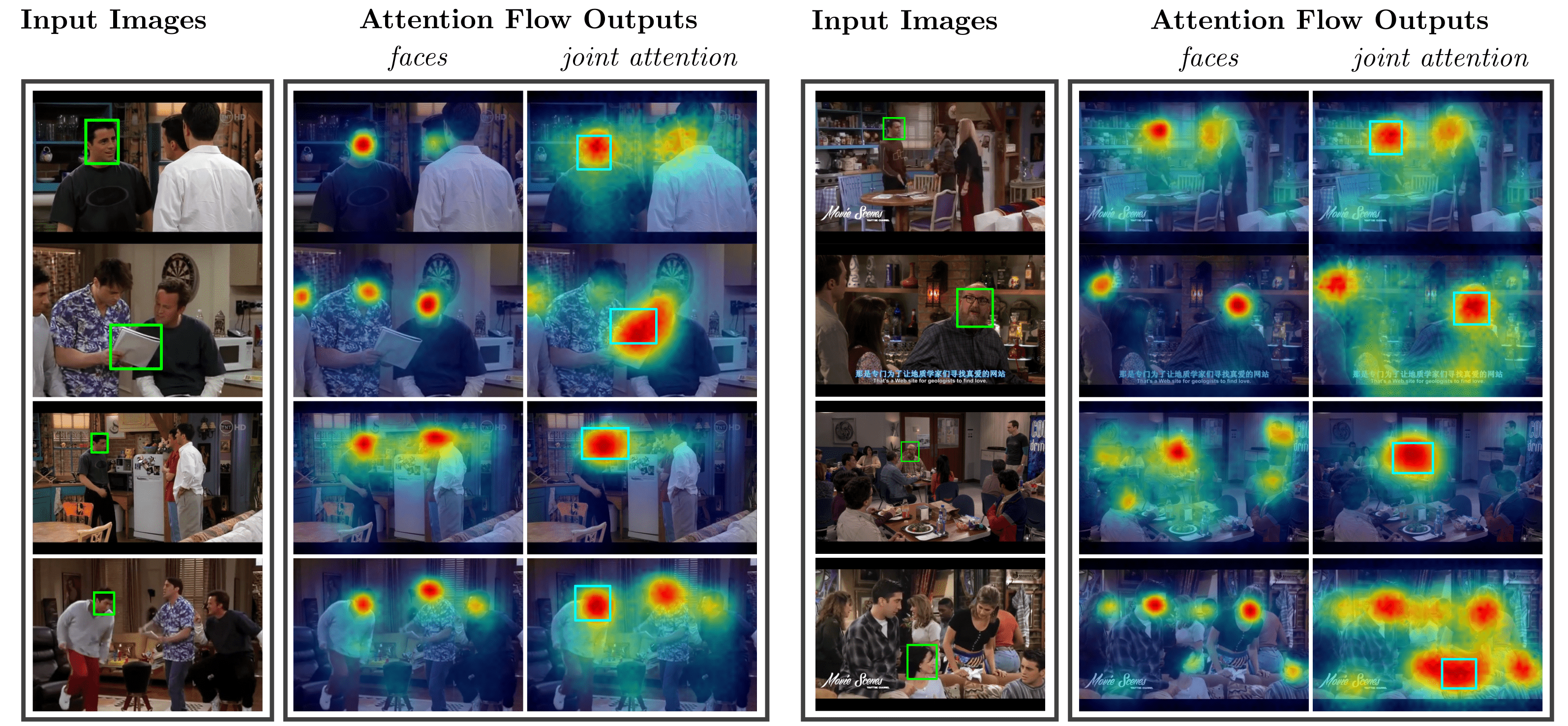}
	\caption{\textbf{Qualitative results of Attention Flow} Bounding boxes on sample test images {\color{green} (green)} show the ground truth attentional focus. The second and third columns are our estimated Attention Flow. In the third column, we also depicted the estimated bounding boxes {\color{cyan} (cyan)}. Figure best viewed in color. }
	\label{figure:05}
\end{figure*}

The output of the dilated residual network that we used as an encoder is 1/8 resolution of the input and its channel size is 2048. The channel-wise attention module, first applies $(4\times 6)$ average pooling, two convolutional layers whose kernel sizes are $3\times 3$, $3\times 2$ with a stride of 2 and 1, respectively. Their channel sizes are 512 and 2048, respectively. The final output is in the size of $C\times 1 \times 1$ and the original encoder output is channel-wise multiplied by these importance weights.  

\begin{table}[b]
	\centering
	\begin{tabular}{ll}
		\toprule
		\textbf{Method} 		&	 \textbf{ $L_2$ distance} \\
		\midrule
		Attention Flow \\ $E_{(lr=0)}$	&  73.77 \\
		$E_{(base\_lr)}$ 	&  70.47 \\
		$E_{(finetune)}$  	&  69.72 \\
		\midrule
		$channel-wise \ attention$  	&  62.84 \\ \midrule
		$spatial \ attention$  		&  65.70 \\
		\bottomrule\\
	\end{tabular}
	\caption{The effect of attention mechanisms in localization of joint attention over the test set of VideoCoAtt dataset.}
	\label{table_attention}
\end{table}

Table \ref{table_attention} shows the results of joint attention localization over the test set of VideoCoAtt dataset. We first tested the following options: To use the encoder as pre-trained features (no learning), to train the encoder in the same learning rate as the generator, and finetuning the encoder by a reduced ($\times 0.1$) learning rate. Channel-wise attention aims to apply feature selection in a learn representation. Thus, we freeze the encoder when training channel-wise attention and generator jointly. This approach reduces the mean $L_2$ distance by 10.92 and 6.88 pixels in comparison with no learning and finetuning, respectively.   

Looking into our spatial attention, we applied spatial attention to the output of the generator in 1/8 resolution ($40\times 60$) before linear upsampling. Spatial attention module takes the estimation of joint attention maps ($2\times H\times W$) and learns a spatial importance on top to better localize the co-attention point. In spatial attention, we use a $3\times 3$ convolutional layer (64), batch normalization, a residual bottleneck module and final convolutional layer to reduce channel size back to 2. Then, a sigmoid activation is applied and the previous predictions are weighted.   

Before using attention mechanisms, we show how accurate we can localize co-attention bounding boxes based on our baseline approach that is depicted in Figure~\ref{figure:02}. After creating saliency guided pseudo-attention maps that we use as the label, our Attention Flow network has two trainable blocks: an encoder, and a generator. The encoder is initialized by ImageNet trained weights. Then, we compared  the following three cases: freeze the encoder and train only generator ($E_{(lr=0)}$), train encoder and generator jointly ($E_{(base\_lr)}$), and learn encoder by transfer learning with a reduced learning rate and train generator from scratch ($E_{(finetune)}$). As it can be seen in Table \ref{table_attention}, transfer learning performs better than the approaches mentioned above and achieves an $L_2$ distance of $69.72$.

Channel-wise attention, which is used between encoder and generator, can predict joint attention with a mean distance of $62.84$, whereas spatial attention after the generator gives $65.70$. Both attention mechanisms improve joint attention localization by $4.02$ and $6.88$ points with respect to our baseline network with encoder fine-tuned in Table \ref{table_attention}. The better performance of our channel-wise attention approach indicates that feature selection on top of deep learning features plays an important role. Weighting features per channel improves their potential as a scene descriptor.

\begin{table}[b] 
	\label{table:results}
	\centering
	\resizebox{\columnwidth}{!}{%
		\begin{tabular}{lll}
			\toprule
			\textbf{Model} 		&	\textbf{Prediction Acc.} &  \textbf{$L_2$ distance} \\
			\midrule
			Random  		& 50.8\%		&  286 \\
			Fixed Bias  	& 52.4\%		&  122 \\
			\midrule
			Gaze Follow \cite{Recasens:2015} 	& 58.7\%		&  102 \\
			Raw Image \cite{Fan:2018} 			& 52.3\%    &  188 \\
			Only Gaze \cite{Fan:2018} 			& 64.0\%    &  108 \\
			Gaze+RP  \cite{Fan:2018} 	    & 68.5\%		&  74 \\
			\midrule
			Gaze+Saliency+LSTM \cite{Fan:2018} 	& 66.2\%    &  71 \\
			Gaze+RP+LSTM  \cite{Fan:2018}	    & 71.4\%	&  \textbf{62} \\
			\midrule
			\textbf{Ours \emph{(w channel-wise att.)}} 	    & \textbf{78.1}\%	&  62.84 \\
			\bottomrule\\
		\end{tabular}
	}
	\caption{\textbf{Quantitative evaluation results} with Prediction Accuracy and \emph{L2} Distance over the test set of VideoCoAtt dataset.}
	\label{table:final_results}
\end{table}

Table \ref{table:final_results} shows our results in comparison with other methods in detection and localization of joint attention. \emph{\textbf{Random}} is acquired by drawing a Gaussian heatmap with a random mean and variance. \emph{\textbf{Fixed Bias}} uses joint attention bias in the TV shows (averaged over the VideoCoAtt dataset) to sample predictions. An alternative to joint attention is to make prediction per person using \emph{\textbf{Gaze Follow}} \cite{Recasens:2015} and combine their attention likelihoods. Other methods are from the reference of VideoCoAtt dataset \cite{Fan:2018} and grouped into two categories: single frame and temporal models. All of these methods \cite{Fan:2018} depend on head detection bounding boxes, region proposal model or saliency estimation even in test time. In terms of used modalities, our approach is similar to their \textbf{\emph{``Raw Image''}} approach. 

Our Attention Flow network with a channel-wise attention detects joint attention with an accuracy of $78.1\%$ over the entire test set of VideoCoAtt. Furthermore, it localizes co-attention bounding boxes with $L_2$ distance of $62.84$. Our method performs significantly better than \cite{Fan:2018}'s single frame with region proposals and gaze estimation. Furthermore, our approach is on par with \textbf{\emph{Gaze+RP+LSTM}} and outperforms it in terms of prediction accuracy by $6.7\%$.

We should note that our model makes this improvement without using any head pose/gaze estimation branch, region proposal maps, and also temporal information. [9]'s models with LSTM leverages 20-30-frame length sequences to improve and smooth prediction performance. We focused on learning an end-to-end model by using only single raw frames. Therefore, as in Table \ref{table:final_results}, our model's performance ($78.1\%$ and $62.84$) is far beyond \cite{Recasens:2015} and \cite{Fan:2018}'s single frame approaches which perform at best, \textbf{\emph{Gaze+RP}}, $68.5\%$ and 74 in joint attention detection and localization, respectively.

Figure \ref{figure:05} depicts the qualitative results of our Attention Flow network on several test images of VideoCoAtt. The ground truth co-attention locations are shown in green rectangles. Estimated face and co-attention likelihoods are overlaid on images. The first channel of our attention maps successfully locates both frontal and side faces. Looking into co-attention estimation, predictions in groups with $3$-$4$ persons are very good. Even though their distances from ground truths are not very large, the last two examples (on the right) are relatively broad. This is due to the difficulty of scenes and a wider angle of view. 

Another point that we should address is the distribution of social formations in the VideoCoAtt dataset. As the dataset is composed of acted scenes mostly from the TV shows, it does not represent the real-world formations such as in learning situations or group work. In addition, when we inspect the failures, we observed that most of them were from complicated cases where many people interest each other. Their faces were far from the camera and difficult to determine their activities (i.e., last two samples on the right side of Figure~\ref{figure:05}). The possible direction in joint attention analysis can be to create training corpus specialized in the desired applications such as group work analysis, therapeutic situations, or children's gaze behaviors. 

\section{Conclusion}\label{ssec:conclusion}
This study addressed a recently proposed problem, inferring joint attention in third person social videos. Our Attention Flow network infers joint attention based on only raw input images. Without using any temporal information and other dependencies such as a face detector or head pose/gaze estimator, we detect and localize joint attention better than the previous approaches. We create pseudo-attention maps by leveraging saliency information to better detect and localize joint attention. Furthermore, we propose two new convolutional attention blocks for feature selection and attention map localization. As inferring joint attention in an end-to-end fashion necessitates a high-level inference, increasing the amount of training data or the network depth will not help. We should note that these attention mechanisms, particularly channel-wise attention blocks for feature selection, are highly essential to select useful features from learned representations and improve localization performance of a heatmap regressor. 

Understanding of joint attention by use of computer vision can help in a wide range of applications such as educational assessment, human-computer interactions, and therapy for attention disorders and as a future work we extend our approach to specialize in these applications and use as a tool for human behavior understanding. 

\paragraph*{Acknowledgements} Ömer Sümer is a doctoral student at the LEAD Graduate School \& Research Network [GSC1028], funded by the Excellence Initiative of the German federal and state governments. This work is also supported by Leibniz-WissenschaftsCampus Tübingen ``Cognitive Interfaces''.

{\small
	\bibliographystyle{ieee}
	\bibliography{egbib}
}

\end{document}